\documentclass{llncs}
\usepackage[utf8]{inputenc}
\usepackage[T1]{fontenc}

\usepackage[ngerman,english]{babel}
\usepackage{graphicx}
\usepackage{hyperref}

\usepackage{comment}
\usepackage{todonotes}
\usepackage{blindtext}

\usepackage{tikz}

\usepackage{wrapfig}
\usepackage{pgfplots,caption}
\pgfplotsset{compat=1.9}

\usepackage{rotating}
\usepackage{pdflscape}
\usepackage{subfig}

\usepackage{url}

\usepackage[shortlabels]{enumitem}

\usepackage{amsmath}

\usepackage{amssymb,array}

\title{Knowledge Representations in\\Technical Systems}
\subtitle{A Taxonomy}

\author{Kristina Scharei, Florian Heidecker, Maarten Bieshaar}

\institute{Intelligent Embedded Systems, University of Kassel, Germany \\
\email{\{kristina.scharei,florian.heidecker,mbieshaar\}@uni-kassel.de}}

\begin{document}

\maketitle

\begin{abstract}
	The recent usage of technical systems 
	in human-centric environments leads to the question, how to teach technical systems, e.g., robots, to understand, learn, and perform tasks desired by the human. Therefore, an accurate representation of knowledge is essential for the system to work as expected. This article mainly gives insight into different knowledge representation techniques and their categorization into various problem domains in artificial intelligence. Additionally, applications of presented knowledge representations are introduced in everyday robotics tasks. By means of the provided taxonomy, the search for a proper knowledge representation technique regarding a specific problem should be facilitated.
\end{abstract}

\section{Introduction}




The increasing number of technical systems (TS) used for services, e.g., robot assistants in homes or factories, results in many challenges to accomplish, e.g., safe human-robot interaction or consistently changing environments for a robot vacuum cleaner in a household. Especially in such cases, TS need to act like a human would. This results in the challenge of how to accomplish this, maybe even better than a human. Therefore, a significant question is: How to represent knowledge required for a specific task such that a TS can understand it and consequently fulfill the required task? In the past years emerged many different knowledge representation (KR) approaches such that the choice of a right KR for a specific task is not straightforward. This article should act as an aid to get an overview of different KR techniques used in everyday applications. The main contributions are:


\begin{itemize}[label=\textbullet]
	\item a taxonomy consisting of different KR techniques for TS 
	\item categorization of previously stated KR techniques in different problem domains of artificial intelligence (AI)
	\item application of various KR techniques in robotic scenarios \textit{activity learning}, \textit{task representation}, and \textit{scene understanding}
	
\end{itemize}

The remainder of this article is as follows:
First, problem domains in AI are presented. Section~\ref{sec:overview} gives an overall overview of possible KR in TS, whereas Petri Nets (PN) as semantic graphs are described in more detail. Section~\ref{sec:application} exemplary combines the presented KR with application scenarios from everyday robotics tasks.

\section{Problem Domains in Artificial Intelligence}
\label{sec:domain}


For the applicability of different KR, it is of advantage to get an overall overview of different challenges in AI first. These challenges are divided into domains, which describe the potential capabilities of an AI technology and type of problem AI can solve \cite{AIDomains}. 
This section will be differentiating various problem domains in AI including some overall application scenarios. Due to varying divisions of problem domains, this section combines multiple sources: \cite{Jain2015}, \cite{Rosenschein1985}, \cite{AIDomains}, and \cite{AIOverview}. 

\subsubsection{Perception} describes knowledge about the environment by transforming raw sensorial inputs into technically processable information. Sensorial inputs can be images, videos, or sound data. \hfill\\
\textit{Possible applications are:} autonomous vehicles, medical diagnosis, or surveillance.

\subsubsection{Reasoning} 
indicates solving problems by means of logical deduction.
This requires knowledge to be already available. \hfill\\
\textit{Possible applications are:} legal assessment, financial asset management, games, or autonomous weapons systems.

\subsubsection{Knowledge} as an AI domain stands for the ability to represent knowledge of the world in an understandable way for TS. Especially, this means dividing knowledge into certain entities, events, or situations which themselves have properties and can even be categorized. \hfill\\
\textit{Possible applications are:} medical diagnosis, drug creation, media recommendation, purchase predictions, or fraud prevention.

\subsubsection{Planning} is the capability of setting and achieving goals. In more detail: a specific future state of the world is desirable, and the sequences of actions to access this state are highly relevant. \hfill\\
\textit{Possible applications are:} logistics, scheduling, navigation, physical and digital network optimization, predictive maintenance, or demand forecasting.

\subsubsection{Communication}
involves the continual updating of mutual knowledge possessed by the speaker and hearer. Key concepts are: understanding language in written and spoken form and communicating with the environment. \hfill\\
\textit{Possible applications are:} voice control, intelligent agents, assistants and customer support, real-time translation of written and spoken languages, or real-time transcriptions.

\subsubsection{Learning} indicates incrementation of knowledge through experience.\hfill\\
\textit{Possible applications are:} workforce learning, intelligent job matching, partially automated assessment.
\hfill\\
\\
Problems in previously mentioned domains can be solved through either statistical methods or traditional symbolic and sub-symbolic methods. 

\subsubsection{Symbolic methods} assume that the world can be reduced to symbol manipulations only. They can be subdivided into two different approaches: logic-based and knowledge-based. \textbf{Logic-based} approaches consist of one out of several logics. They are often used in case of problem-solving. \textbf{Knowledge-based} approaches, on the other hand, are based on ontologies and huge databases of notions, information, and rules. 

\subsubsection{Statistical methods} solve specific sub-problems using mathematical tools. There are two different approaches regarding statistical methods: probabilistic methods and machine learning techniques. \textbf{Probabilistic methods} allow acting in incomplete information scenarios. \textbf{Machine learning} tools allow TS to learn from given data.

\subsubsection{Sub-symbolic methods} contain no explicit knowledge. Embodied Intelligence and Search/Optimization techniques are included in these methods. \textbf{Embodied Intelligence} assumes that higher intelligence requires a body, which can consist of, for instance, movement, interaction, or visualization functions. \textbf{Search and optimization} allow intelligent examination of the search area which can lead to one of many possible solutions. The solution found with these techniques is not always globally optimal, but "good enough" to work further with the results. \textbf{Machine learning} can be counted to these methods as well because the knowledge in these tools is almost always sub-symbolic.
\hfill\\

\begin{landscape}
\begin{figure} 
	\centering 
	\noindent\includegraphics[width=\linewidth,height=\textheight,keepaspectratio]{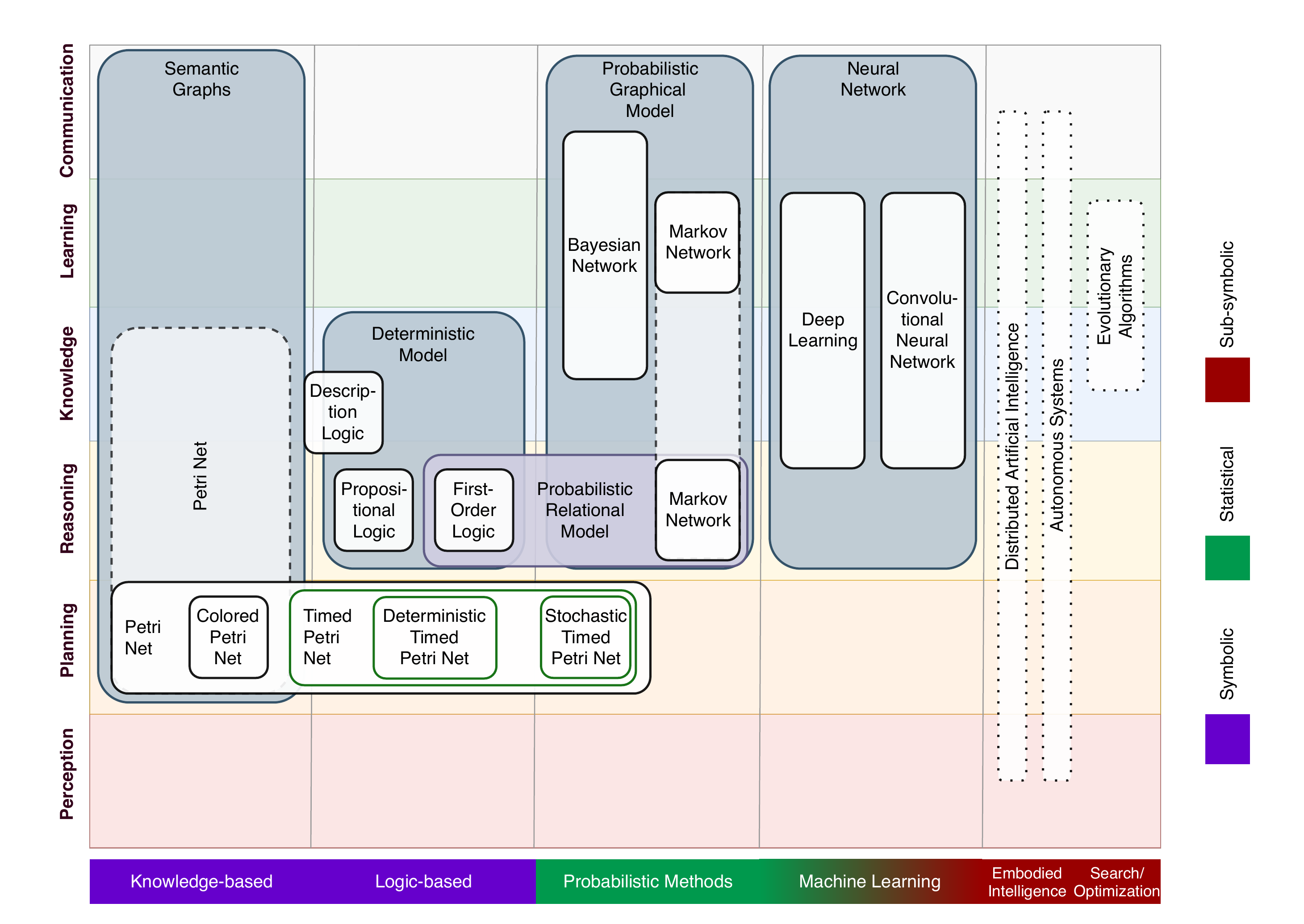} 
	\caption{Overview of different knowledge representations (inspired by \cite{AIDomains})} 
	\label{fig:overview} 
\end{figure} 
\end{landscape}

Figure \ref{fig:overview} displays the presented partitioning from above. The y-axis displays the problem domains in AI; the x-axis shows the different methods to achieve results in these problem domains. Within the grid are displayed different KR techniques. The big gray rectangles represent the four central representations dealt with in this article. Rectangles with dotted lines are possible representations in the sub-symbolic area but will not be further discussed here. Dashed lines represent areas that can be used by the KR as well but are also not further discussed. The reason for the classification being, as shown in Fig.~\ref{fig:overview}, will be understandable at the end of the article.
\hfill\\

The next section is focusing on symbolic and statistical methods (gray boxes in Fig.~\ref{fig:overview}), providing different KR techniques in the form of graphs and networks.

\section{Overview: Graph and Network Representations}
\label{sec:overview}

Graphs are one key concept in computer sciences, which are often used to model real-world systems. A \textit{graph} is an ordered pair of sets $(V,E)$ with $V$ being a set of \textit{vertices} (or \textit{nodes}) and $E$ a set of \textit{edges} connecting vertices. The fact that they are universally applicable is what makes them so popular. Some of the qualities when using graphs:  They are easy to search for paths, rapid while adding and deleting knowledge or quick and easy regarding visualization \cite{Kudase2016}. 
Networks build a subgroup of graphs because a network can be defined as a graph.

As graphs are universally applicable many KR are based on them, which will be presented in the next section after a quick introduction into deterministic models.

\subsection{Deterministic Models}
\label{subsec:deterministicModels}

In a deterministic model, 
states are determined by the model's parameters as well as the previous states.
Given a set of initial conditions, the produced output of a deterministic model will always be the same.
To express knowledge in a verifiable way, different logic-approaches are presented in consistent formal theories, which allow the deduction of new provably correct knowledge from existing knowledge by means of systematic or syntactic manipulations of logical propositions \cite{Nyga2017phd}.

The next sections will be focusing on different common logic approaches following \cite{Nyga2017phd}.

\subsubsection{Propositional Logic}
Propositional logic (PL) is one of the most basic formalisms in logic-based knowledge representation. \textit{Sentences} in PL consist of atomic propositions from an alphabet of symbols $\Sigma$, which can take values true ($\top$) or false ($\bot$). An \textit{interpretation} assigns a truth value to every atomic proposition. It is possible to construct more complex sentences by applying \textit{logical junctors} on simple sentences. 
Assumption: $\phi$ and $\psi$ are sentences in PL. New sentences can be constructed by \textit{negation} ($\neg\phi$), \textit{conjunction} ($\phi\wedge\psi$), \textit{disjunction} ($\phi\lor\psi$), \textit{implication} ($\phi\to\psi$), and \textit{bi-implication} ($\phi\leftrightarrow\psi$). 
Atomic and complex statements are also called \textit{logical formulae}.
An interpretation $I$ of statement $\phi$ is called \textit{model of $\phi$} if the statement is true under interpretation $I$ ($I \models \phi$).
Essential characteristics of PL are:

\begin{itemize}[label=\textbullet]
	\item \textbf{Satisfiablity:} A formula $\phi$ is satisfiable, if and only if (in the following written as \textit{iff}) there is at least one interpretation under which $\phi$ is true. Otherwise, it is called unsatisfiable.
	\item \textbf{Validity:} A formula $\phi$ is valid, iff $\phi$ is true under all interpretations.
	\item \textbf{Entailment:} A formula $\phi$ entails another formula $\psi$ ($\phi\models\psi$), iff every model of $\phi$ is also a model of $\psi$.
\end{itemize}
PL is \textit{decidable}. It states that there is an algorithm that can, for each formula, decide whether a property is fulfilled or not.

\subsubsection{First-Order Logic}

Due to the limited possibilities to express statements, first-order logic (FOL) extends PL by introducing constant, predicate, function, and variable symbols. \textit{Constant symbols} represent entities or attributes in the world, \textit{predicate symbols} denote relations between these entities. A \textit{term} addresses an entity with syntactic expressions, whereas every constant and variable symbol is a term. A \textit{function} can be applied to terms. Terms without variables are called \textit{ground terms}.

The most powerful tool in FOL is the usability of universal ($\forall$) or existential ($\exists$) quantifiers. They extend the validity of a sentence to hold for \textit{all} (in case of a universal quantifier) or \textit{at least one} (in case of an existential quantifier) substitutions of the referenced variable in the sentence.

FOL, in general, is only \textit{semi-decidable}. This means that there is an algorithm that can for each formula decide if a property is fulfilled, but cannot decide whether the property is not fulfilled, which can result in an infinite loop. 
Still, restricted variants of FOL have been developed, which restore full decidability.

\subsubsection{Description Logic} \label{subsubsec:DL}(DL) is a decidable subset of FOL which allows representing aspects of the real world regarding upper ontologies of concepts and instances of it. DL is used in semantic web or robotics, e.g., KnowRob system \cite{KnowRob}. A \textit{knowledge base} (or ontology) in DL consists of terminological and assertional axioms.
 \textit{Terminological axioms} define which categories of objects there are in the respective domain, which relations (e.g., specialization and generalization) between these categories hold, and what kind of properties the objects may have. \textit{Assertional axioms} describe a specific state of affairs of an application domain in terms of concepts and roles from the terminological axioms.
If $A$ and $B$ are concepts, new concepts can be defined as follows:

\begin{itemize} [label=\textbullet]
	\item \textbf{definition $A \doteq B$:} entities of A are entities of B and vice versa
	\item \textbf{intersection $A \sqcap B$:} entities of A and B at the same time
	\item \textbf{union $A \sqcup B$:} entities of A or B
	\item \textbf{complement $\neg A$:} entities of all types but A
	\item \textbf{inclusion $A \sqsubseteq B$:} all entities of type A are also of type B, but the opposite direction is not necessarily true
\end{itemize}




Though deterministic models are useful regarding problem-solving and powerful for representing and reasoning about knowledge, a significant disadvantage of these models is the lack of probability or uncertainty in general. Therefore, the application of such models is limited.

\subsection{Probabilistic Graphical Models} 
In this section, particular emphasis is placed on the approach that the environment of a TS cannot be described in simple rules (as assumed in Sect.~\ref{subsec:deterministicModels}) and therefore cannot be modeled perfectly. Likelihoods and uncertainty dominate the knowledge in this representation and can be used in manipulation recognition and understanding \cite{Paulius2018}. Two often-used probabilistic graphical models (PGM) will be explained further in the next sections, following mainly \cite{Nyga2017phd} and \cite{Paulius2018}.

\subsubsection{Bayesian Network}
\begin{figure}
	\vspace{-12pt}
	\centering
	\includegraphics[scale=0.2]{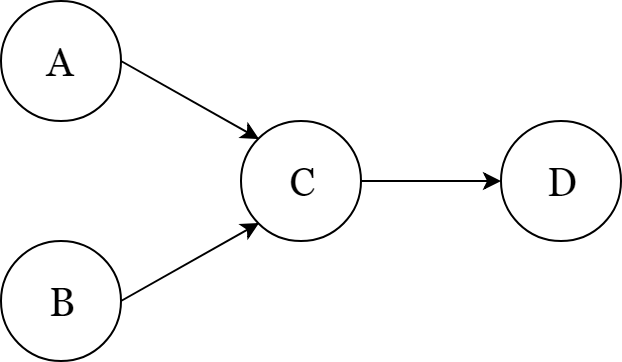}
	\caption{Exemplary Bayesian network with four nodes A, B, C, D.}
	\label{fig:BN}
\end{figure}

Bayesian Networks (BN) are directed and acyclic graphs that provide a formalism to specify dependencies regarding causal relationships of variables. Nodes within the graph represent entities or concepts; edges describe probabilistic dependencies between nodes. A \textit{parent} is a node with at least one outgoing edge. $Par(X_{i})$ is a set which represents the parents of a node $X_{i}$ on which $X_{i}$ depends. Figure~\ref{fig:BN} shows an exemplary BN consisting of four nodes (A, B, C, D) where each node represents a random variable. In this example, $Par(X_{i})$ with , e.g., $X_{i}=C$ is equal to $\{A,B\}$.
The existence of a path between two nodes indicates an influence between them. Otherwise, the nodes, which follow a local \textit{Markov assumption}\footnote{\textbf{Markov assumption}: The future is independent of the past given the present.}, are conditionally independent. Due to this characteristics, BNs can be used for making inferences about unobserved variables and additionally determine more complex relationships.

The probability of unknown discrete variables can be calculated with Eq.~\ref{eq:likelihood}, called the \textit{Bayes' theorem}. The joint distribution over variables $X_{1},...,X_{n}$ is given by Eq.~\ref{eq:jointDistributionBN}. According to Eq.~\ref{eq:jointDistributionBN}, the joint distribution of the exemplary network in Fig.~\ref{fig:BN} is: $P(A,B,C,D) = P(A) * P(B) * P(C|A,B) * P(D|C)$

\begin{equation}
\label{eq:likelihood}
P(X|Y) = \frac{P(Y, X)}{P(Y)} = 
\frac{P(Y|X)P(X)}{P(Y)}
\end{equation}

\begin{equation}
P(X_{1},...,X_{n}) = \prod_{i=1}^{n} P(X_{i}|Par(X_{i}))
\label{eq:jointDistributionBN}
\end{equation}

Instead of using purely discrete variables it is possible to apply continuous variables to a BN called \textit{Linear Continuous Gaussian} BN (LCG-BN). As the name indicates, continuous variables follow a linear Gaussian distribution. A network of this type can therefore represent probabilities using a Gaussian distribution.

LCG-BNs are only one variation of simple BNs which exist. There are many more.

\subsubsection{Markov Network}

As a trade-off between the generality of full joint distributions and restricted BN structures, Markov Networks (MN) represent another family of PGMs.
Unlike the BN a MN, also known as Markow Random Field (MRF), has no directed edges, which denotes that there is no directional flow of influence and therefore can capture cyclic dependencies. An equal flow of influence between connected nodes is the outcome. Similar to BNs, a local Markov assumption denotes conditional dependence. Variables in MNs do not always follow probability distributions. Instead, a parametrization by means of \textit{factors} (functions representing the relationship) is applied to each \textit{clique} (subset of nodes where all nodes are connected).

The joint distribution is a Gibbs distribution given by: 

\begin{equation}
P(X=x)=\frac{1}{Z}\prod_{c \in G}^{}\psi_{c}(x_{|c|})
\label{eq:jointDistributionMN}
\end{equation}
where $c$ indicates cliques in a network $G$ for a specific world $x$, $\psi_{c}$ is a factor, and $x_{|c|}$ is a projection of the values of all variables in $x$ that are part of $c$. 
Consequently, the joint distribution relies on all factors, not only a single factor.

\begin{equation}
Z = \sum_{x' \in X}^{} \prod_{c \in G}^{} \psi_{c}(x'_{|c|})
\label{eq:normalizeMN}
\end{equation}
 Equation~\ref{eq:normalizeMN} shows the \textit{partition function} $Z$ which normalizes the probability masses of all possible worlds to get a proper probability distribution.

\begin{figure}
	\vspace{-12pt}
	\centering
	\includegraphics[scale=0.25]{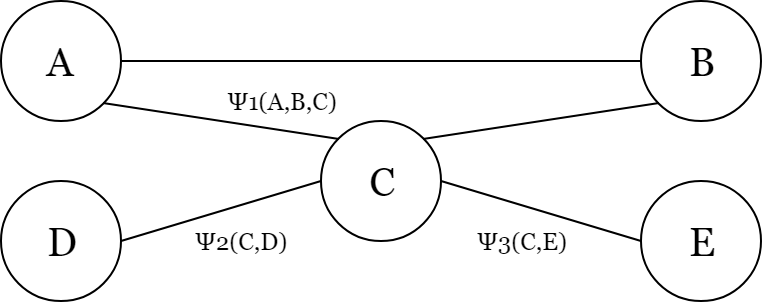}
	\caption{Exemplary Markov network consisting of five nodes A, B, C, D, E and maximal cliques \{A,B,C\}, \{C,D\} and \{C,E\} including their clique potentials $\psi_{1}$, $\psi_{2}$ and $\psi_{3}$ (based on \cite{Nyga2017phd}).}
	\label{fig:MN}
\end{figure}

Figure~\ref{fig:MN} shows an exemplary Markov network consisting of three cliques. The joint distribution given by Eq.~\ref{eq:jointDistributionMN} is then calculated by:\hfill\\
$P(A,B,C,D,E) = \psi_{1}(A,B,C) * \psi_{2}(C,D) * \psi_{3}(C,E)$

MNs are typically used in representing transitions of nodes (sometimes called \textit{states}), which are not necessarily irreversible.

\textit{Conditional Random Fields} (CRF) are special instances of MNs which encode conditional distributions. Given a set of observed variables $X$ 
and target variables $Y$,  
a CRF models a distribution
based on $X$, i.e. $P(Y|X)$. 
The distribution does not consider cliques, which not only consider observable variables.
They are typically used for modeling sequential data, where the past influences present and future decisions. Observed variables are utilized for detemining the outcome of hidden state variables.

\subsection{Probabilistic Relational Model}

PGMs are graph-based structures that are useful to recognize independencies between variables. Therefore, a compact representation and more efficient computations are enabled. Due to the possibility to simply interpret graphs, domain experts and knowledge engineers can design PGMs straightforwardly. Despite the advantages, a significant drawback is the fixed number of variables which have to be known beforehand. Sometimes, though, it is desired to support varying numbers of random variables.
The aim of probabilistic relational models (PRM) is to overcome limitations of logics as well as PGMs by using KR formalisms like FOL and further involve uncertainty and inconsistency. Markov Logic Networks (MLN) are the most widely used PRMs which combine the expressiveness of FOL and probabilistic semantics of MN \cite{Nyga2017phd}.

Bayesian Logic Networks can be used as well but are not further discussed in the following section. For more information, please refer to \cite{Jain2011BLN}.

\subsubsection{Markov Logic Network}

A MLN comprises a set of FOL formulas $F_{i}$ and real-valued weights $w_{i}$ with regard to the $i$-th formula, each written as $\langle F_{i}, w_{i} \rangle$. $w_{i}$ determines the correctness of a formula: Large weights denote important formulae, small weights imply less relevant formulae. Unlike in FOL, most MLNs contain \textit{typed predicate arguments}, i.e., all arguments of a predicate are bound to a set of values whereas in FOL predicate arguments are in one universal set \cite{Nyga2017phd}.

\begin{figure}
	\centering
	\includegraphics[scale=0.05]{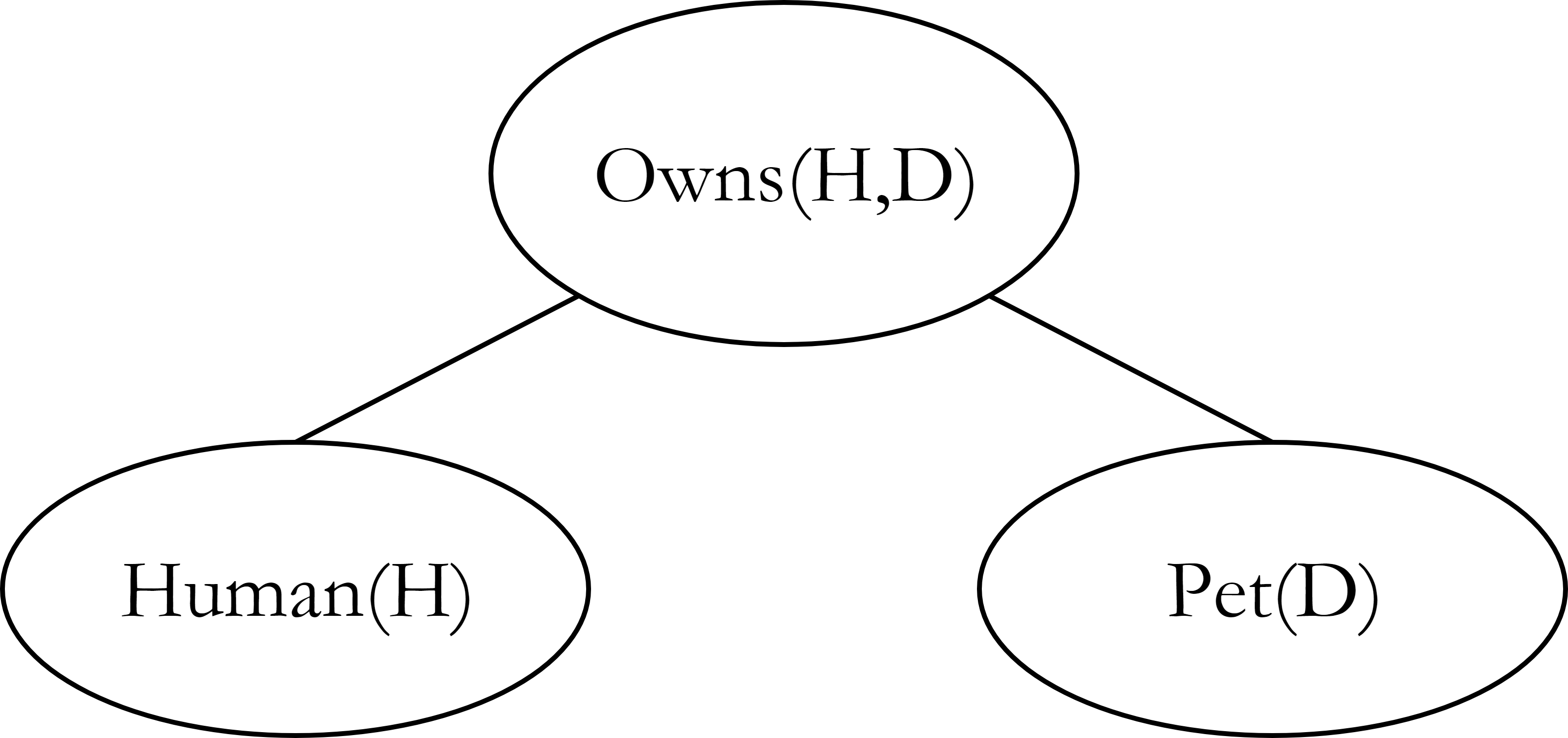}
	\caption{Exemplary Markov Logic Network representing the FOL statement that a human H can own a pet D ($\forall H\  \forall D\  Owns(H,D) \to Human(H) \leftrightarrow Pet(D)$) (based on \cite{Paulius2018})}
	\label{fig:MLN}
\end{figure}

Figure~\ref{fig:MLN} shows an exemplary MLN.
 If there is a human "Ryan" owning dog "Wilfred" it follows that Owns(Ryan, Wilfred), Human(Ryan)
 and Pet(Wilfred). \cite{Paulius2018}.
The simplicity of MLN makes them quite general and powerful yet intuitive regarding representation languages for uncertain knowledge \cite{Nyga2017phd}.

\subsection{Artificial Neural Networks}

By means of probabilistic models, it is possible to describe scenarios including uncertainty, which is getting closer to the real-world scenario than, e.g., deterministic models. One of the goals in TS is not only to represent the world as it is but to even act in the world like, e.g., a human would. 

Therefore, it is desirable to use the concept of machine learning to represent knowledge and furthermore allow TS to learn from knowledge that is already available \cite{Paulius2018}.

This section introduces the concept of Artificial Neural Networks (NN), which are mostly used in research, following mainly \cite{BioInspiredAI}
\vspace{7pt}

\begin{figure}
	\centering
	\includegraphics[scale=0.4]{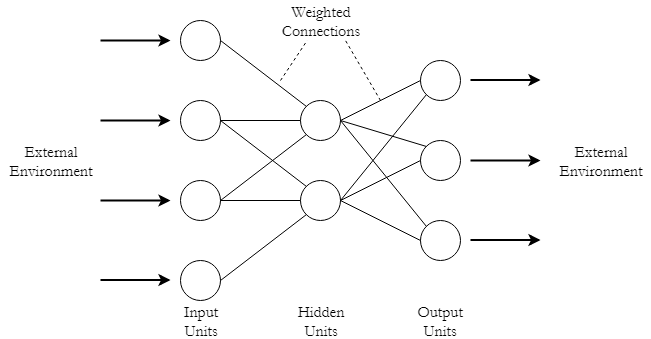}
	\caption{Exemplary Neural Network with one hidden layer (based on \cite{BioInspiredAI})}
	\label{fig:NN}
\end{figure}

Inspired by biology, NNs attempt to capture the behavioral and adaptive features of biological nervous systems in computational models.
It consists of several interconnected units whereas \textit{input units} receive information from the environment, \textit{output units} affect the environment, and \textit{hidden units} communicate only within the network. A different view on NN is to look at the \textit{layers} of a NN, whereas one layer consists of one or more units. Each unit implements an operation that gets active if the total incoming signal is larger then its threshold. After the activation of a unit a signal is sent to all units to which the activated unit is connected. The connection between the units multiplies the signal by a signed weight. The result of a NN to an incoming signal from the environment depends on the architecture and the connection strengths within it. Figure~\ref{fig:NN} shows an exemplary NN including one hidden layer, which is not \textit{fully connected} (all units between two layers are connected). It should be clear that the knowledge within the network is distributed across the NNs connections.


NN learn by modification of weights when presented
with stimulation from the environment. Usually, learning requires
several repeated presentations of the set of input patterns. There are several
types of learning rules, each displaying specific functionalities and are applicable
to specific architectures. Typically, all synaptic connections within the
NN change according to the same learning rule. Characteristics of NNs are:

\begin{itemize}[label=\textbullet]
	\item \textbf{Robustness:} to various types of signal degradation,
	such as input noise or malfunctioning of connection.
	\item \textbf{Flexibility:} NNs can be applied to several types of problems. 
	However, not every type of NN can be applied to any type of problem.
	\item \textbf{Generalization:} A trained NN can provide a correct response to a not yet used input pattern, which is similar to the training data.
	The ability of the network to generalize
	the response to a new pattern depends on the extent to which the new
	pattern can be described by the learned invariant features.
	This characteristic of NNs is useful since at times it is difficult to model all possible situations a system may be exposed to.
	\item \textbf{Content-based retrieval:} NNs retrieve memories by matching
	contents and can do so even when the input patterns are incomplete or
	corrupted by noise. More familiar patterns are recognized faster
	than items that are different or seen less frequently. 
\end{itemize}

\subsubsection{Convolutional Neural Networks}

Convolutional Neural Networks (CNN) were developed to drastically reduce the number of variables to train by using \textit{weight sharing}. Thereby, CNNs are more effective compared to densely connected NNs. CNNs are easy to implement in spite of the uncertainty due to the number of free parameters. In particular, learning concepts from data of different types and modalities in robotics displays a promising application area \cite{Paulius2018}.

\subsubsection{Deep Learning}


Deep neural networks are often utilized as feature detectors or classifiers. By storing them, they can be reused to, e.g., detect affordances (cf. Sect.~\ref{sec:application}). With a properly trained network, a technical system can even detect such details for unknown objects, which is made possible through extensive training. \cite{Paulius2018}


Also, a combination of both, namely a \textit{Deep Convolutional Neural Network} (D-CNN), is possible.
\hfill\\

However, one major drawback of utilizing NNs is the requirement of much training data. For example, in image recognition
thousands of images are required to learn or detect objects in scenes. Additionally, it can be challenging to interpret and explain how each layer segregates data \cite{Paulius2018}.

\subsection{Semantic Graphs}

Semantic graphs (SG) are another subset of graphs whose nodes and edges describe semantic concepts and details between entities, as observed in demonstrations. For example, spatial concepts can be described with nodes being objects. Edges between them denote similarities regarding position.
In association with SG, it is interesting to mention the term \textit{ontology} and its connection to SG. An ontology is a "formal description of knowledge as a set of concepts within a domain and the relationships that hold between them" \cite{Ontologies} (cf.~Sect.\ref{subsubsec:DL}). Ontologies provide shareable and reusable KR but also enable adding new knowledge about the domain. Once relationships between concepts are provided in an ontology, automated reasoning about data is enabled and beyond that such reasoning can be implemented with SGs. In other words, SGs use ontologies as their semantic schemata. \cite{Ontologies} 

This chapter will be focusing on Petri Nets (PN) but there are many more SGs (including \textit{semantic event chains} (SEC) \cite{Aksoy2011}, \textit{parse graphs} \cite{Zhu2015}, etc.) available, which will not be examined any further.

\subsection{Petri Nets}
\label{subsec:PN}

This section introduces the concept of PNs following \cite{Wang2007PetriNF} unless otherwise specified.
PN combine a
well-defined mathematical theory with a graphical representation of the dynamic behavior of systems. They have been used to model event-driven systems such as computer networks, communication systems, etc. A PN is a bipartite directed graph and consists of four objects: \textit{places}, \textit{transitions}, \textit{directed arcs} and \textit{tokens}. Directed arcs connect places to transitions or transitions to places. 



Formally, a PN is a five-tuple $N = (P, T, I, O, M_{0})$ with the characteristics shown on the left side beneath. On the right side, the formal characteristics are applied to the simple PN shown in Fig.~\ref{fig:petriNet} (a).

\begin{tabular}[t]{@{}>{\raggedright\arraybackslash}p{0.65\textwidth}}
\begin{enumerate}[(1)] 
	\setlength{\itemsep}{3pt}
	\item \textbf{$P = \{p_{1},...,p_{m}\}$} is a finite set of places.
	\item \textbf{$T = \{t_{1},...t_{m}\}$} is a finite set of transitions.
	\item \textbf{$I: T \times P \rightarrow N$} is an input matrix that   \ \ specifies directed arcs from places to transitions.
	\item \textbf{$O: T \times P \rightarrow N$} is an output matrix that specifies directed arcs from transitions to places.
	\item \textbf{$M_{0}: P \rightarrow N$} is the initial marking.
\end{enumerate}
\end{tabular}
\begin{tabular}[t]{@{}>{\raggedright\arraybackslash}p{0.35\textwidth}@{}}
\begin{enumerate}[(1)] 
	\setlength{\itemsep}{2pt}
	\item$P = \{p_{1}, p_{2}, p_{3}, p_{4}\}$
	\item$T = \{t_{1}, t_{2}, t_{3}\}$
	\item$I = \begin{pmatrix} 
				2 & 0 & 0 & 0 \\
				0 & 1 & 0 & 0 \\
				0 & 0 & 1 & 0 \\
				\end{pmatrix}$
	\item $O = \begin{pmatrix} 
				0 & 2 & 1 & 0 \\
				0 & 0 & 0 & 1 \\
				0 & 0 & 0 & 1 \\
				\end{pmatrix}$
	\item$M_{0} = (2\ 0\ 0\ 0)^{T}$.
\end{enumerate}
\end{tabular}

\subsubsection{Enabling Rule}A transition $t$ is \textit{enabled} if each input place $p$ of $t$ contains at least the
number of tokens equal to the weight of the directed arc connecting $p$ to $t$, i.e., $M(p)\geq I(t, p)$ for all $p$ in $P$. 
\subsubsection{Firing Rule}Only enabled transitions can \textit{fire}. The firing of an enabled transition $t$ removes the number of tokens equal to $I(t, p)$ from
each input place $p$, and deposits in each output place $p$ the
number of tokens equal to $O(t, p)$.
The new marking is:
$M'(p) = M(p) - I(t, p) + O(t, p)$ for all $p$ in $P$. \hfill\\

\begin{figure}%
	\vspace{-18pt}
	\centering
	\subfloat[Simple Petri Net with four places $p_{1}, p_{2}, p_{3}, p_{4} $ and transitions $t_{1}, t_{2}, t_{3}$]{{\includegraphics[width=5.5cm]{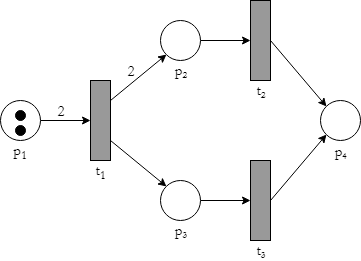} }}%
	\qquad
	\subfloat[Firing of transition $t_{1}$]{{\includegraphics[width=5.5cm]{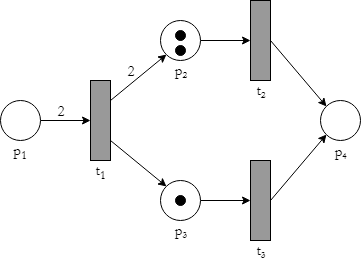} }}%
	\caption{Exemplary Petri Net and how it works (based on \cite{Wang2007PetriNF})}%
	\label{fig:petriNet}%
	\vspace{-18pt}
\end{figure}

Consider again the simple PN shown in Figure~\ref{fig:petriNet} (a). Under initial marking $M_{0}$ only $t_{1}$ is
enabled. According to the firing rule, firing of $t_{1}$ results in a new marking $M_{1} = (0\ 2\ 1\ 0)^{T}$.
The new token distribution of this PN is shown in Figure \ref{fig:petriNet} (b). 


\textit{Sequential execution}, \textit{conflicts}, \textit{concurrency}, \textit{synchronization}, \textit{mutually exlusiveness}, \textit{priorities} and \textit{resource constraints} can be modeled effectively with PN.

Important properties of a PN are:

\begin{itemize}[label=\textbullet]
	\item \textbf{Reachability: } Marking $M_{j}$ is said to be \textit{reachable}
		from a marking $M_{i}$ if there exists a sequence of transition firings that transforms $M_{i}$ to $M_{j}$. It is \textit{immediately reachable} if $M_{i}$ directly results in $M_{j}$.
	\item \textbf{Safeness: } Place $p$ is said to be $k$-bounded if the number of tokens in $p$ is always less than or equal to $k$ ($k$ is a
		non-negative integer number) for every marking $M$ reachable from the initial marking $M_{0}$, i.e., $M \in R(M_{0})$.
		It is safe if it is 1-bounded.
		A Petri net $N =(P, T, I, O, M_{0})$ is $k$-bounded (safe) if each place in $P$ is $k$-bounded (safe). It is
		unbounded if $k$ is infinitely large.
	\item \textbf{Liveness: }According to \cite{Murata1989}, a transition $t$ in a PN is:\\
		(1) \textit{L0-live / dead}: there is no firing sequence in $L(M_{0})$ in which $t$ can fire.\\
		(2) \textit{L1-live / potentially firable}: $t$ can be fired at least once in some firing sequence in $L(M_{0})$.\\
		(3) \textit{L2-live}: $t$ can be fired at least $k$ times in some firing sequence in $L(M_{0})$ given any positive integer $k$.\\
		(4) \textit{L3-live}: $t$ can be fired infinitely often in some firing sequence in $L(M_{0})$.\\
		(5) \textit{L4-live / or live}: $t$ is L1-live in every marking in $R(M_{0})$.
\end{itemize}

With regard to representing knowledge, PNs are often used to display processes or sequences of actions \cite{Paulius2018}.

\subsubsection{Colored Petri Nets}

A significant disadvantage of simple PN is the production of large and unstructured specifications for the systems being modeled. To tackle the issue, high-level PNs were developed to allow compact system representations with, e.g., Colored Petri Net (CPN). Each token in a CPN has a color which indicates the identity of the token and each place and transition has attached a set of colors. 
With respect to a color, transitions can fire. By firing, tokens are deleted from the input places and therefore added to output places as in original PN. In contrast to this, there is a dependency between the color of the transition firing and the token. A transition firing may change the color.


This type of PN can be used for, e.g., generating colored \textit{petri net plans} (PNP), which are a formal model for representing and executing multi-robot plans
\cite{Steccanella2016ColouredPN}.

\subsubsection{Timed Petri Nets}

Since time is a central component in dynamic systems and often one does not want to neglect the time component when modeling it, the usage of time variables or constraints can be beneficial. PNs can as well contain such time variables. This subgroup of PN is then called a \textit{timed petri net} (TPN). The definition of a TPN consists of three specifications:

\begin{itemize}[label=\textbullet]
	\item \textbf{topological structure}: same as in standard PN
	\item \textbf{labeling of the structure}: assigning numerical values to one or more of the following objects: transitions, places and arcs connecting places and transitions
	\item \textbf{firing rules}: depend on the PN and its time variables
\end{itemize}

\cite{Chao2012} showed that TPN can be used for, e.g., \textit{multimodal interaction modeling}.

Deterministic and stochastic timed PNs are the most known of this type and will be explained in the next sections.

\subsubsection{Deterministic Timed Petri Nets}

A deterministic timed Petri Net (DTPN) is a six-tuple $(P,T,I,O,M_{0},\tau)$ where $(P,T,I,O,M_{0})$ is a PN, $\tau: T \to R^{+}$ is a function associating transitions with deterministic time delays.

A transition $t_{i}$ in a DPTN can fire at a time $\tau$ iff:
\begin{itemize}[label=\textbullet]
	\item for any input place $p$, there is a number of tokens equal to the weight of the directed arc connecting $p$ to $t_{i}$ in the input place continuously for time interval $[\tau - \tau _{i}, \tau]$, with $\tau _{i}$ being the associated firing time of transition $t_{i}$.
	\item After transition firing, each of its output places $p$ receive the number of tokens equal to the weight of the directed arc connecting $t_{i}$ to $p$ at time $\tau$.
\end{itemize}

\subsubsection{Stochastic Timed Petri Nets}

Stochastic timed Petri nets (STPN) are PN where stoachstic firing times are linked to transitions. Mainly, it builds a stochastic process.
They are a graphical model and offer great convenience to a modeler in arriving at a credible, high-level model of a system. The most straightforward choice for the individual distributions of transition firing times is the negative exponential distribution. 
Due to the memoryless property of this distribution, the stochastic process is a continuous time-homogeneous Markov chain. Its state space is in one to one correspondences with marking in $R(M_{0})$, set of all reachable markings.

\section{Applications of Knowledge Representation Techniques in Robotics}
\label{sec:application}

\cite{Paulius2018} investigated different application scenarios in robotics and scanned several research articles on how knowledge was represented there. This section sums up their realizations and displays three different application scenarios from robotics and how the KR from Section~\ref{sec:overview} can be used to solve the given problems. 
As an additional contribution, the application scenarios are classified into problem domains from Section~\ref{sec:domain} according to the applications found in the given research articles and therefore can be assigned to Fig.~\ref{fig:overview}.

\subsection{Activity Learning}

To guide a learning process in robotics, researchers often refer to the concept of \textit{affordance learning}. \textit{Affordance} is specified as "qualities or properties of an object that define its possible uses or make clear how it can or should be used" \cite{Affordances}. 
They are useful for determining likely actions and effects attributed to a set of tools or objects available to a robot for manipulation. 
\textit{Affordance learning} focuses on "how objects respond to actions which act upon them" \cite{Paulius2018}. An advantage of this learning method upon \textit{reinforcement learning} (get rewarded for right actions; punished for wrong actions) \cite{Sutton2018} is the generation of a deeper understanding.\\
\textbf{Categorization in Problem Domains:} \textit{reasoning}, \textit{learning}, \textit{knowledge}



\subsubsection{Using Bayesian Networks to represent Affordances}

Affordances can be represented with probabilistic models like BN or MN.

 \cite{Moldovan2012} used LCG-BNs for representing the continuous state of objects as they respond to actions done upon them. Exceptional performance was shown in predicting action-effects.

\cite{Sun2013} describe the likelihood of object interaction by setting up a BN. This can be learned by means of human behavior observation.
The produced models were shown to improve the accuracy of other tasks in activity recognition or object classification and can be used to teach robots how to perform tasks using paired objects.

\cite{Jain2013} applied BNs to learn about the effects of actions on objects by means of demonstrations with tools. Their primary objective was to identify functional features, i.e., geometric features relevant to a tool's function. By means of inference with the BN, a prediction of tool effects unknown to the robot is possible.



\subsubsection{Using Markov Networks to represent Affordances}

\hfill\\
\cite{Koppula2012} used MRFs for describing relationships between activities and objects within the scene. The purpose of the generated model was to understand full-body activities occurring in videos that they successfully demonstrated.

\cite{Kjellstrom2011} used CRFs to perform both object classification and action recognition for hand manipulations. A good performance could be achieved in classifying objects, actions occurring, and both simultaneously.



\cite{Zhu2014} used a MLN as knowledge base for inferring object affordances in activities. Therefore, they collected information of the objects in the form of features. 
Due to FOL statements, the network could be easily queried about objects affordances. They used them to predict a pose of a human while performing tasks using particular objects.


\subsubsection{Using Semantic Graphs to represent Affordances}
\hfill\\
\cite{Aksoy2011} created semantic graphs after segmentation. They focused on understanding the relationship between objects and
hands in the environment. Also, they wanted to generalize graphs for representing activities and identify future instances
of these events. The encoding of objects is based on manipulations only, which they store in matrices called \textit{semantic event chains} (SEC). Similar events can be recognized based on the SECs. A great advantage of this work above others is that similarity even can be measured with different objects, orientations, hand positions, etc.

\cite{Sridhar2008} separated scenes into blobs using segmentation as well. 
Afterward, they clustered the blobs as a semantic graph based on their use in videos for affordance detection. The type of semantic graphs they used were \textit{activity graphs}, which are structures describing the spatial and temporal relationships in a single video.

\cite{Zhu2015} learned affordances by identifying probable usage of tools. They use a \textit{spatial-temporal parse graph} to store these concepts.
Further information about parse graphs can be found in \cite{Xiong2016} and \cite{Tu2013}.
By using inference, further knowledge can be extracted.


\subsubsection{Using Machine Learning Techniques to represent Affordances}
Aside from learning affordances with graphical representations, researchers also use machine learning
methods for activity representations, e.g., Support Vector Machines (SVM) \cite{Ugur2012}.

\subsection{Task Planning}

In this section, different approaches to map the higher-level knowledge to a robot's understanding in order to perform tasks are discussed. In terms of task KR, the represented knowledge should include low-level features for a robot's understanding as well maintain a level such that it is interpretable by a human.
Any type of knowledge can be constrained to an ontology and therefore describe the working space of the robot in a format that is also interpretable by humans. 
As discussed in Sect.~\ref{subsec:PN} SGs are most suitable for this kind of application scenario.
\\
\textbf{Categorization in Problem Domains:} \textit{planning}


\subsubsection{Using Semantic Graphs for Task Representations}

\hfill\\
\cite{Galindo2008} provided a semantic map-based representation that integrates \textit{causal knowledge} as well as \textit{world knowledge}. Causal knowledge describes how actions affect the state of the robot's environment; world knowledge displays what the robot
knows about objects within the world.
Their map uses DL for combining factual knowledge about the environment, e.g., position or geometry of objects, 
with general semantic knowledge about the domain. 

\cite{Ramirez2017} 
developed a robot transfer learning system. They created SGs such that it includes skills for executing kitchen tasks.

\cite{Costelha2007,Costelha2012} used PNs for representing robot tasks. They chose PNs over, e.g., Finite State Automata, due to the potential of representing concurrent system action and sharing of resources. With regard to the structure of PNs, they filtered out specific plans which can not happen or should be avoided. 
As a result, they create PNPs which are mostly a combination of ordinary actions and sensing
actions using control operators.


\subsection{Scene Understanding}


In addition to robot learning in form of understanding and recognizing activities, a lot of research has branched off into understanding the entire scene. 
Therefore, an understanding of \textit{semantic concepts}, e.g., location of objects, and \textit{spatial features}, e.g., orientation, is required.
\\
\textbf{Categorization in Problem Domains:} \textit{knowledge}, \textit{learning}, \textit{communication}

\subsubsection{Using Deep Learning Techniques to represent Scene Understanding}


\cite{Lenz2013} utilized deep learning to identify grasping points on objects.
Their work showed that this approach can also be extended to objects which have never been seen
before, even while these objects may still be in a clutter.



\cite{Levine2016} approached to learn grasps and to perfect hand-eye coordination with the use of monocular camera images, a large number of robot arms, and a D-CNN. Inputs of the D-CNN includes two images, namely one of the current scene and the original setting.

\cite{Yang2015} learned manipulation action plans by investigating videos from the internet. Two CNNs were required to do so: The first network recognizes objects and grasping points. The second network generates actions regarding robot manipulation.



\subsubsection{Using Bayesian Networks to represent Scene Understanding}
\hfill\\
\cite{Kollar2013} developed an interactive system such that a robot can interpret commands and at the same time it is feasible to find out what the robot knows. The utilized probabilistic model understands natural language such that the robot can interact with a human. This way, further concepts can be learned.



\subsubsection{Using Semantic Graphs to represent Scene Understanding}
\hfill\\
\cite{Bastianelli2013} developed a system for building a semantic map, including world knowledge. A human can communicate with the robot and enforce him to update his knowledge. 
They use the \textit{task of simultaneous localization and mapping} (SLAM) to obtain a 2D
map of the environment.

\subsection{Results}

The main outcomes of Sect.~\ref{sec:application} are as follows:
\begin{itemize}[label=\textbullet]
	\item It is striking that SGs are used to represent knowledge in all three scenarios. This shows that they are universally applicable, with PN being one of them.
	\item Additionally, the example scenarios show that for each problem there are many different possible ways to represent knowledge. Sometimes the right choice is being made with particular attention to details; sometimes it is simply a design issue.
\end{itemize} 

As mentioned in Sect.~\ref{sec:domain} finally a categorization of the different used KR techniques into problem domains in AI is performed. Therefore, the categories of the different application scenarios presented in Sect.~\ref{sec:application} are used to assign the KR techniques providing a solution for the scenario to the problem domains, resulting in Fig.~\ref{fig:overview}

It should be clear that this categorization only refers to the scientific works presented in this article. Inevitably, there are more possible uses of the introduced KRs as indicated in Fig.~\ref{fig:overview}.

\section{Conclusion and Outlook}

This article presents different possible KR techniques for usages in TS. Therefore, a classification of the various techniques into AI domains was performed. Additionally, several application scenarios from the robotics area were presented and realizations of it shown. Due to the large diversity of KR techniques, considering the information structure was shown to be necessary. Each of the presented techniques is suitable for at least one kind of problem but can be inappropriate in another scenario. The created taxonomy provides an assistance of when to use each KR technique and aims to facilitate the choice of a right KR. Obviously, this taxonomy only gives a rough overview and finding a good-enough KR is a huge topic and not straightforward.

Promising approaches can be seen in storing \textit{information in clouds} such that each TS can gather this information through a network. Similar to this \textit{collaborative knowledge}, "learners’ joint activities to acquire or create new knowledge" \cite{Collaborative}, could be an interesting view on KR. Either way, finding a proper KR still will be a problem that has to be solved.

\section*{\large Acknowledgment}

This work results from the project DeCoInt$^2$, supported by the German Research Foundation (DFG) within the priority program SPP 1835: "Kooperativ interagierende Automobile", grant number SI 674/11-1.

\bibliographystyle{apalike}
\bibliography{aBib}

\end{document}